\pdfoutput=1
\documentclass[12pt]{article}

\usepackage{style/ACL2023}

\usepackage{times}
\usepackage{latexsym}

\usepackage[T1]{fontenc}

\usepackage[utf8]{inputenc}

\usepackage{microtype}

\usepackage{inconsolata}

\title{Faithful Chart Summarization with ChaTS-Pi}

\author{
Syrine Krichene,
Francesco Piccinno,
Fangyu Liu,
Julian Martin Eisenschlos\\
Google DeepMind, Z{\"u}rich \\
\texttt{\{syrinekrichene,liufangyu,piccinno,eisenjulian\}@google.com}
}

\usepackage{amsthm}
\usepackage{framed}
\usepackage{mdwlist}
\usepackage{colortbl}
\usepackage{xcolor}
\usepackage{nicefrac}
\usepackage{booktabs}
\usepackage{amsfonts}
\usepackage[T1]{fontenc}
\usepackage{bold-extra}
\usepackage{amsmath}
\usepackage{amssymb}
\usepackage{bm}
\usepackage{graphicx}
\usepackage{mathtools}
\usepackage{multirow}
\usepackage{multicol}
\usepackage[normalem]{ulem}
\usepackage{lipsum}
\usepackage{float}
\usepackage{ifthen}
\usepackage[ruled,vlined]{algorithm2e}
\usepackage{subcaption}
\usepackage{tabularx}
\usepackage[utf8]{inputenc}
\usepackage[english]{babel}
\usepackage{amsfonts} 
\usepackage{cleveref}
\usepackage{bbm}
\usepackage{listings}

\usepackage{latexsym,comment}
\usepackage{tikz}
\usepackage{xspace}
\usepackage{xr}
\usetikzlibrary{shapes,arrows}

\usepackage{enumitem}
\setlist[itemize]{align=parleft,left=0pt..1em,itemsep=0pt}

\newcommand{\gem}[1]{\mbox{\textsc{gem}}}

\newcommand{\hidetext}[1]{}
\newcommand{\ignore}[1]{}

\newif\ifcomment\commenttrue

\newif\ifinfotabs\infotabsfalse

\newif\ifsubscripterror\subscripterrortrue

\ifcomment
\newcommand{\pinaforecomment}[3]{\colorbox{#1}{\parbox{.8\linewidth}{#2: #3}}}
\else
\newcommand{\pinaforecomment}[3]{}
\fi

\ifinfotabs
\newcommand{\infotabstext}[1]{#1}
\else
\newcommand{\infotabstext}[1]{}
\fi

\ifsubscripterror

\else

\fi

\newcommand{\smallurl}[1]{ \begin{tiny}\url{#1}\end{tiny}}

\definecolor{lightblue}{HTML}{3cc7ea}
\definecolor{CUgold}{HTML}{CFB87C}
\definecolor{grey}{rgb}{0.95,0.95,0.95}
\definecolor{ceil}{rgb}{0.57, 0.63, 0.81}
\definecolor{UMDred}{HTML}{ed1c24}
\definecolor{UMDyellow}{HTML}{ffc20e}
\definecolor{darkgreen}{HTML}{008f00}

\newcommand\sota{state-of-the-art\xspace}

\newcommand{\metric}{\textsc{ChaTS-Critic}\xspace}
\newcommand{\pipeline}{\textsc{ChaTS-Pi}\xspace}

\usepackage{scalerel}  %
\usepackage{svg}

\newcommand{\desiredheight}{0.36cm}

\def\criticemoji{\resizebox{!}{\desiredheight}{\includesvg{figures/detective.svg}}}
\def\piemoji{\resizebox{!}{\desiredheight}{\includesvg{figures/shortcake.svg}}}

\usepackage{amssymb}%
\usepackage{pifont}%
\newcommand{\cmark}{\ding{51}}%
\newcommand{\xmark}{\ding{55}}%

\theoremstyle{definition}

\makeatletter
\newcommand*{\addFileDependency}[1]{%
  \typeout{(#1)}
  \@addtofilelist{#1}
  \IfFileExists{#1}{}{\typeout{No file #1.}}
}
\makeatother

\begin{document}
\maketitle

\begin{abstract}
    Chart-to-summary generation can help explore data, communicate insights, and help the visually impaired people.
Multi-modal generative models have been used to produce fluent summaries, but they can suffer from factual and perceptual errors.
In this work we present \metric{} \criticemoji, a reference-free chart summarization metric for scoring faithfulness.
\metric{} is composed of an image-to-text model to recover the table from a chart, and a tabular entailment model applied to score the summary sentence by sentence.
We find that \metric{} evaluates the summary quality according to human ratings better than reference-based metrics, either learned or n-gram based, and can be further used to fix candidate summaries by removing not supported sentences.
We then introduce \pipeline{} \piemoji, a chart-to-summary pipeline that leverages \metric{} during inference to fix and rank sampled candidates from any chart-summarization model.
We evaluate \pipeline{} and \metric using human raters, establishing state-of-the-art results on two popular chart-to-summary datasets.\footnote{Code and demo at \href{https://hf.co/spaces/chats-pi/chats-pi}{hf.co/spaces/chats-pi/chats-pi}.} 

\end{abstract}

\section{Introduction}
Chart summarization requires faithfully extracting quantitative data and describing them using natural language. Recent natural language generation (NLG) studies have explored different flavors of chart-to-summary generation tasks including caption generation for scientific figures \cite{hsu-etal-2021-scicap-generating}, chart summary generation \cite{kantharaj-etal-2022-chart}, or analytical textual descriptions for charts \cite{zhu-etal-2021-autochart}. These tasks can be advantageous for the visually impaired \citep{benetech-making-graphs-accessible} as well as for automating interpreting complex domains such as finance data-analysis, news reporting, and scientific domains \citep{siegel2016figureseer}.

\begin{figure}[!t]
    \centering
    \includegraphics[width=1.\linewidth]{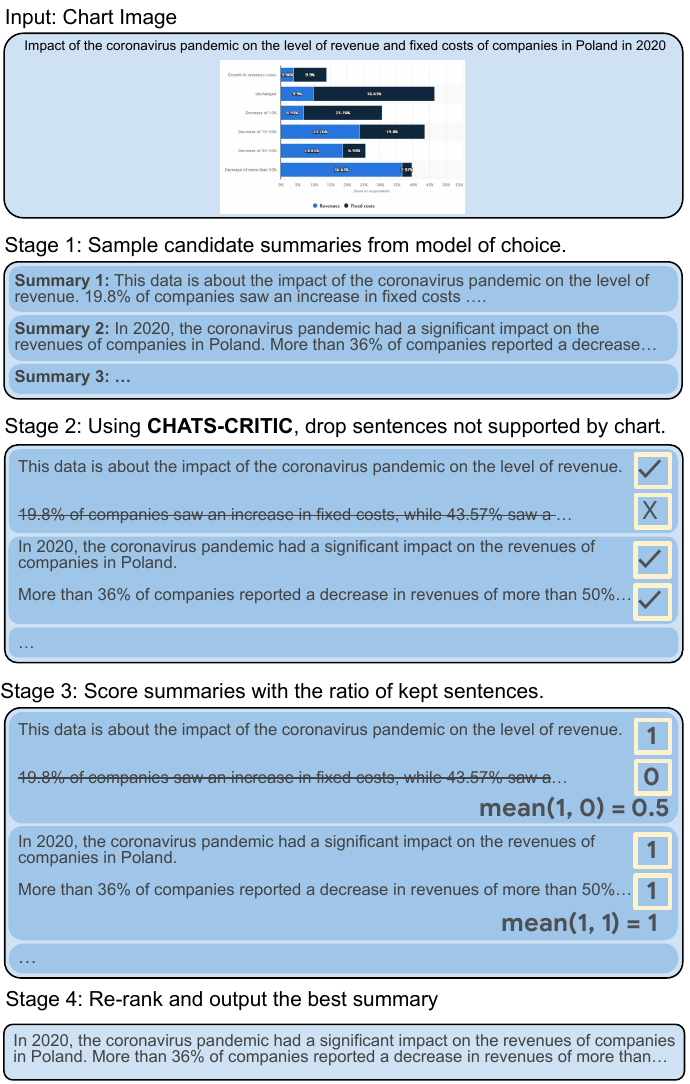}
    \caption{\pipeline generates multiple summaries given the chart using any summarization model. Each summary is then \emph{repaired} by dropping refuted sentences according to the \metric{} sentence scoring. 
    Finally, we rank the summaries by computing the ratio of sentences that were kept.
    }
    \label{fig:ChaTS-Pi}
    \vspace{-2.0ex}
\end{figure}

While a wide range of models and techniques have been applied for chart summarization, faithfulness remains a major challenge for the task.
Specifically, the models often misread details in the charts (due to perceptual mistakes) or miscalculate the aggregations (due to reasoning flaws).
To overcome some of these limitations, \emph{optical character recognition} (OCR) models and object detection systems are usually employed to extract meta-data such as axis, values, titles, legend~\cite{luo2021chartocr,masry-etal-2022-chartqa}. These data are then used as auxiliary inputs to finetune NLG models. Nonetheless, these modeling efforts are still limited by two fundamental issues (i) training \& evaluation dataset quality and (ii) the reference-based metrics being used for evaluation.
As examples, two widely used datasets, Chart-to-Text~\cite{kantharaj-etal-2022-chart} and SciCap~\cite{hsu-etal-2021-scicap-generating}, are automatically extracted from web articles and academic journals. As a result, the summary references are prone to \emph{hallucination}, i.e. the reference might contain context that cannot be entailed solely by the chart content. Training on this data can encourage the NLG models to improvise/hallucinate. Besides, the auto-extracted summaries sometimes emphasize only certain aspects of the chart, missing out critical insights.

\begin{figure}[!ht]
    \centering
    \includegraphics[width=1.0\linewidth]{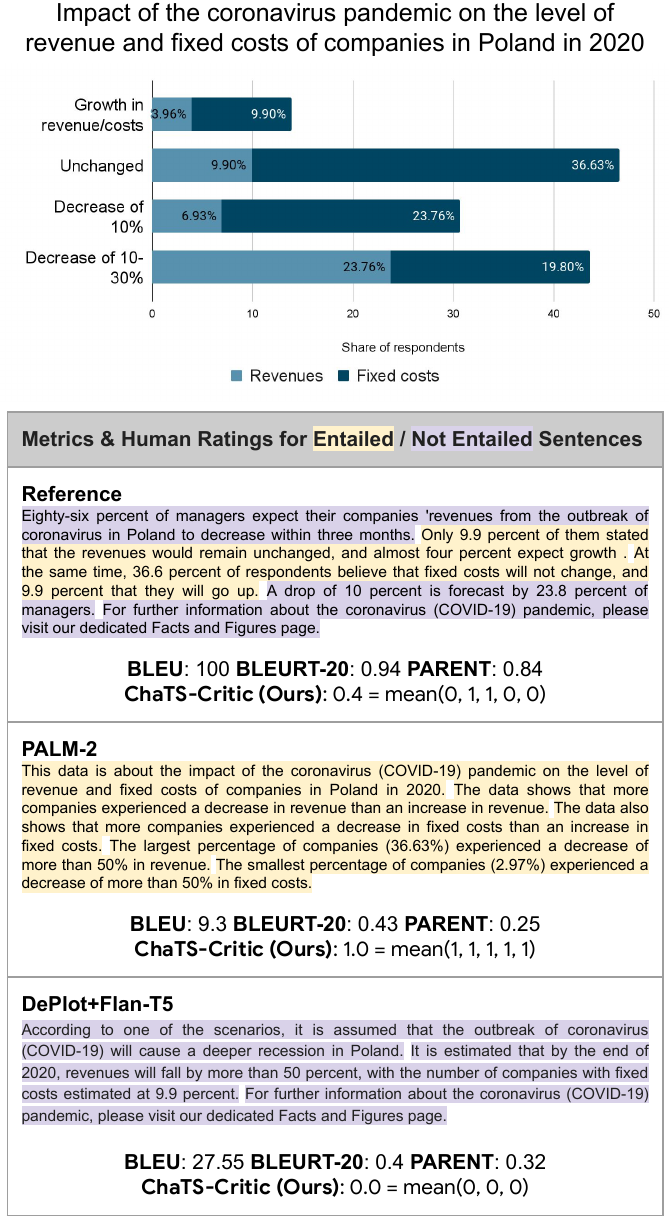}
    \vspace{-4ex}
    \caption{This example from \citet{kantharaj-etal-2022-chart} showcases the limits of reference-based metrics for summary evaluation: (1) the reference text often contains extra information that is not present in the chart which skews the evaluation, and (2) the reference-based metrics can fail at capturing unreferenced but correct sentences. In comparison, \metric{} better reflects the human ratings for summary faithfulness.
    }
    \label{fig:example_statista}
    \vspace{-3ex}
\end{figure}

On the other hand, n-gram based metrics such as BLEU~\cite{papineni-etal-2002-bleu}, or learned metrics such as BLEURT~\cite{sellam-etal-2020-bleurt} rely only on gold references. They are not capable of recognizing 
unreferenced but correct insights since they solely rely on the reference for scoring the summaries, 
as shown in \Cref{fig:example_statista}. This issue is especially pronounced when the gold references are noisy, which is the case for Chart-to-Text and SciCap.
Last but not least, reference-based metrics also heavily penalize summary style mismatches, giving an artificial disadvantage to LLMs which are not tuned on the task data ~\cite{maynez-etal-2023-benchmarking}.

This motivates building a reference-free critic \metric{} (\Cref{fig:CHATS-CRITIC}) that can be used as a metric
to score and re-rank summaries.
We additionally introduce \pipeline{} (\Cref{fig:ChaTS-Pi}) that leverage \metric{} scores to generate a high quality summaries.
We summarize our contributions as follows:

\begin{figure}[!ht]
 \vspace{-1.0ex}
    \centering
    \includegraphics[width=0.90\linewidth]{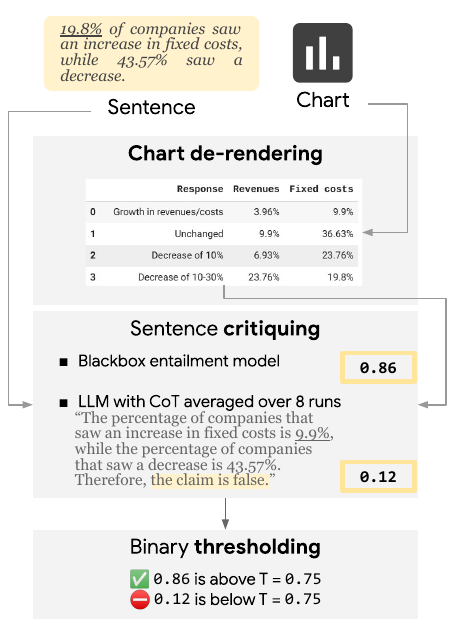}
  \vspace{-2.0ex}
    \caption{\metric{} is composed of a de-rendering model to extract the table from the chart, and a table entailment model. The latter can be a blackbox table entailment model (e.g., TabFact as benchmarked in \Cref{tab:chats-critic-size}) or an LLM; in latter case, we use CoT prompt and average over~$8$ samples. In the figure, the threshold to reach a binary decision is set to $T=0.75$. The chart icon refers to the same plot of \Cref{fig:example_statista}.}
    \label{fig:CHATS-CRITIC}
    \vspace{-2.0ex}
\end{figure}

\begin{enumerate}[noitemsep]
    \item We present \metric{}, a reference-free metric composed of a model that extracts the underlying table data from the chart and a table-entailment model acting on each sentence within a chart summary.
    \item We design \pipeline{}, a pipeline that (i) generates multiple candidate summaries using a generative model, either fine-tuned or with in-context learning; (ii) then leverages \metric{} to refine the summaries by dropping unsupported sentences; (iii) computes a summary score to rank the summaries by penalizing summaries with dropped sentences to increase the fluency, and (iv) outputs the best one.
    \item To assess the efficacy of \metric{}, we juxtapose human preferences against both \metric{} and other prevailing metrics. Our results indicate that \metric{} aligns more consistently with human evaluations. Furthermore, when contrasting \pipeline{} with other leading models that serve as baselines, \pipeline{} establishes \sota on two popular English benchmarks.

\end{enumerate}

\section{The \metric{} \criticemoji{}~ metric}
As shown in \Cref{fig:CHATS-CRITIC}, \metric{} is composed of a chart de-rendering model that generates the table content of the input chart image, and a table entailment model applied on a sentence level.
This motivation stems from the observation that fine-grained evaluations are simpler than full-summary evaluations, mirroring the ease observed in human assessments~\cite{krishna-etal-2023-longeval}.%

\paragraph{Chart de-rendering.} To utilize the information in chart, previous works have incorporated a step to transcribe the image across modalities to a data table~\citep{luo2021chartocr,kantharaj-etal-2022-chart,liu-etal-2023-deplot}. This process of \emph{de-rendering} enables leveraging downstream text model capabilities to process the information, rather than relying on an image model, which is typically only pre-trained on natural images.
Similarly, in our work we start with a de-rendering step to extract the table $t$ from an image of a chart $C$~\cite{liu-etal-2023-deplot}.\footnote{We also tested end-to-end models using chart images as direct input, but the current de-rendering-based pipeline yielded the best performance.}

\paragraph{Sentence level faithfulness score $f(s)$} (interchangeably referred to as \metric{}) is a sentence-level score defined as the probability of entailment of a sentence $s$ conditioned on the de-rendered table $t$.
This can be accomplished using a fine-tuned table-specialized model such as TAPEX~\cite{liu2022tapex} and TAPAS-CS~\cite{eisenschlos-etal-2020-understanding}, or by prompting an LLM such as PALM-2~\cite{anil2023palm}.
For the latter case, we can use few-shot examples with chain-of-thought as well as ensemble across $K$ model runs by averaging the binary scores produced in each run, to improve the entailment accuracy, as shown in the example in \Cref{fig:CHATS-CRITIC}. The sampling produces a similar effect as Monte-Carlo dropout used by \citet{steen-etal-2023-little}.

\section{The \pipeline{}~\piemoji{}~ pipeline}
\pipeline{} \piemoji, presented in in Figure~\ref{fig:ChaTS-Pi}, shorthand for \underline{Cha}rt-\underline{T}o-\underline{S}ummary \underline{Pi}peline , generates a set of candidate summaries in stage 1, then uses \metric{}'s per sentence scores to repair (stage 2) score the summaries (stage 3) and re-rank them (stage 4).
This is done by removing sentences with low entailment scores and picking the candidate summary with the highest \emph{Summary-level faithfulness score}.%

\paragraph{Summary-level faithfulness score $F(S)$} is a per summary score defined as the ratio of kept sentences:
  \vspace{-3ex}
\[
F(S) = \frac{1}{|S|}\sum_{i=1}^{|S|} \mathbbm{1}_{[T, 1]}\left( f(s_i) \right) %
\vspace{-1ex}
\]
 
where $\mathbbm{1}_{[T, 1]}(f(s_i))$ is the indicator function with $T$ as the threshold, which is equal to $1$ if $f(s_i) > T$ and 0 otherwise.

\section{Experimental Setup}
\label{sec:experiment_setup}

We assess our methods on diverse datasets to prove their broad applicability.

\subsection{Datasets}
\paragraph{Chart-to-Text~\cite{kantharaj-etal-2022-chart}} 
is a large-scale benchmark for chart summarization including bar, line, area, scatter and pie charts, composed of two data sources: \emph{Statista} (35k examples) and \emph{Pew Research} (9k).\footnote{\href{https://statista.com}{statista.com} and \href{https://pewresearch.org}{pewresearch.org}}

\paragraph{SciCap~\cite{hsu-etal-2021-scicap-generating}} is a large-scale benchmark for figure-captioning. It is extracted from science arXiv papers published between 2010 and 2020 and contains more than 2 million figures. We use 3 subsets of SciCap: \emph{First Sentence collection} (133k), \emph{Single-sentence Caption} (94k data points), and \emph{Caption with No More than 100 Words} (131k).

\paragraph{TabFact~\cite{chen2019tabfact}} is a large-scale dataset for table-based fact verification. It contains $16$k Wikipedia tables as evidence for $118$k human-annotated statements. This dataset allows us to study fact verification with semi-structured inputs. We use it to evaluate the entailment accuracy of \metric.

All our models are developed on the dev sets of the mentioned benchmarks and performances are reported on their test sets. We include more detailed descriptions and processing details of the benchmarks in \Cref{app:datasets}.

\subsection{Setups for evaluation \& comparison}

\paragraph{Evaluating \metric.} We evaluate the quality of \metric by comparing the model output entailment to human annotated examples randomly extracted from the Chart-To-Text (Statista). We also evaluate the metric's correlation with human judgments on summary level. We compare \metric to reference-based metrics, including BLEU~\cite{papineni-etal-2002-bleu}, PARENT~\cite{dhingra-etal-2019-handling} that takes the table into account to compute n-gram similarity and as well as BLEURT-20~\cite{sellam-etal-2020-bleurt, pu-etal-2021-learning}, a learned metric.

\paragraph{Evaluating \pipeline.} We report a wide range of metrics' scores across the three benchmarks. We compare \pipeline applied on different base models, as well as  state-of-the-art baselines in the literature which do not rely on \pipeline where applicable. 
The SOTA baselines include PaLI \citep{chen2023pali} and MATCHA \citep{liu-etal-2023-matcha} MATCHA on Chart-To-Text; M4C-Captioner \citep{horawalavithana2023scitune} on SciCap. We additionally train and evaluate PaLI \citep{chen2023pali} ourselves to report more comprehensive results across different benchmarks and metrics. 
All metrics are reported on the sentence level except for the correlation study.

\subsection{Our models}
\label{subsec:models}

\paragraph{Plot-to-table model.} As described, our approach relies on a plot-to-table translation model. For all our models, we make use of DePlot~\cite{liu-etal-2023-deplot}, a \sota model for extracting table contents from chart images (i.e. chart de-rendering).\footnote{More details about DePlot can be found in \Cref{app:Deplot}.} The de-rendered table is passed to a generative text-to-text model for further processing.

\paragraph{Generative models.}
We use two models for summary generation with the de-derendered table from last step as input.
We adapt a FLAN-T5~\cite{suresh2023semantic} base model with table embeddings to enhance table structure understanding, following the scheme of TabT5~\cite{andrejczuk-etal-2022-table}.  We fine-tune this model for each datasets for 220k training steps with a batch size of 128. We denote this setup as DePlot+FLAN-T5 (see \Cref{app:flan_t5}). 
The second approach is PALM-2 (L)~\cite{anil2023palm} with in-context learning. The full prompt is described in \Cref{app:LLM-prompting}. %
We experiment with other models including end-to-end models in \Cref{app:Evaluation_of_the_summary_repair_pipeline}.

\paragraph{\metric{}} is used for the \pipeline{} pipeline and as an additional metric in our experiments.
We experiment with different model sizes and families for \metric{}'s entailment component. 
When not specified, \metric{} uses DePlot and PaLM-2 (L) with Chain-of-thought~\citep{wei2022chain} for the entailment model (shown in \Cref{fig:CHATS-CRITIC}).
The full prompt is reported in \Cref{app:prompting}.

\section{Results}
\subsection{Meta evaluation of \metric}
\label{subsec:evaluation-setup}

\metric{} is evaluated by assessing its correlation with human ratings and the overall quality of the generated summaries.
We randomly sampled $60$ different charts from Chart-To-Text (Statista) test set and surveyed the \emph{Entailment}, \emph{Relevance}, and \emph{Grammaticality} (see \Cref{app:annotation-framework}) on the sentence and summary level when appropriate, making a multidimensional quality metric~\cite{huang-etal-2020-achieved}.
The provenance of the summaries is hidden to prevent biasing the raters. The raters are $10$ volunteering researchers from our institution (not including the authors).
We refined the guidelines with a small sample of examples and raters before formally starting the survey. The Cohen’s Kappa in this sample between pairs of raters is 0.61, which suggests substantial agreement \citep{landis1977measurement}. In the formal survey, the raters annotated the full set, one rater per example. 
As shown in \Cref{fig_app:annotation-example} in \Cref{app:annotation-framework}, we display the chart alongside the title, then for each sentence we ask the rater if is (1) \emph{entailed}, (2) \emph{relevant}, and (3) \emph{grammatically correct}.
The full annotation guidelines are reported in \Cref{app:annotation_guidelines}. We collect annotations for four data collections presented in \cref{tab:results_Human_annotation_size}. 
\begin{table}[ht]
\centering
\vspace{-0.5ex}
\resizebox{0.96\linewidth}{!}{
\begin{tabular}{lcc}
    \toprule
    \textbf{Human annotation set (60 charts)} & \textbf{Avg \# sentences} & \textbf{Avg \# not entailed} \\
    \midrule
    Reference                                       & $2.5$ & $1.0$ \\
    PALM-2                                          & $4.3$ & $1.1$ \\
    \piemoji(PALM-2, \criticemoji(PALM-2))          & $5.4$ & $1.0$ \\
    \piemoji(PALM-2, \criticemoji(DePlot, PALM-2))  & $5.0$ & $0.5$ \\
    \bottomrule
\end{tabular}
}
\caption{Human annotation data collections sizes. \piemoji: \pipeline generates a set of 10 candidates using PALM-2, and \criticemoji: \metric considers either the original tables or the generated ones using DePLot.}
\label{tab:results_Human_annotation_size} 
\end{table}
In the two collections using \pipeline, the predictions are generated without dropping the unsupported sentences, to allow a thorough analysis of \metric quality.

\paragraph{Entailment performance.}
We compare \metric{} against a no-op baseline $f(x) = 1$, where no sentences are filtered, as binary classifiers acting on each sentence.
As such, we report Accuracy, F1 and AUC in \Cref{tab:results_Human_annotation}.
AUC measures the ability of a binary classifier to distinguish between the classes $\{0,1\}$. When no sentences are dropped, no classifier is used, the AUC is equal to $50\%$. 
In this use case we can focus on F1 results to evaluate the quality of the input sentences. 
Precision and recall are reported to showcase the trade-off of \metric{} at a threshold of 0.75. 
We show that \metric{} significantly improves upon all the metrics reaching better Precision-Recall trade-off.

The reference summaries in SciCap are extracted automatically, implying that extra information might be present that cannot directly be deduced from the provided chart and metadata alone. As expected, the F1 score is low when considering all sentences entailed (i.e. baseline $f(x) = 1$). Our proposed metric improves F1 by $11$ points and increases AUC by $31.5$ points. For the three other datasets, the summaries' quality is already better than the reference. Thus, the gain is less significant: by $1$ to $2$ points for F1 and $20$ to $22$ for AUC.

We report the Pearson coefficient and the p-value in \Cref{tab:results_Human_annotation}. For all the sets, the p-value is significantly small, indicating a high probability of observing a correlation to human ratings. The Pearson coefficient indicates that \metric has a human rating correlation from moderate ($>30$) to strong ($>50$). %

\begin{table*}
\centering
\vspace{-1ex}
\resizebox{.98\linewidth}{!}{
\begin{tabular}{lllllllll}
\toprule
    \textbf{Annotation set} & \textbf{Sentence Selection}& \textbf{Acc.}& \textbf{Balanced Acc.} & \textbf{Recall} & \textbf{Precision} & \textbf{F1}    &\textbf{AUC}     & \textbf{Pearson (p-value)}\\
    \hline %
    \multirow{3}{*}{Reference}          & $f(x) = 1$              &  $60.0$ & $50.0$          & $100.0$          & $60.0$           & $75.0$          & $ 50.0$         &   $ -- ( --)$          \\
                                        & \criticemoji                 &  $82.0$ & $79.45$           & $92.22$          &$80.58$           & $\textbf{86.01}$& $\textbf{81.56}$&   $62.2( 1.9e-17)$   \\
    \hline
    \multirow{3}{*}{PALM-2}             & $f(x) = 1$              &  $75.48$ & $50.0$          & $100.0$          & $75.48$          & $86.03$         & $ 50.0$         &   $ --(  --)$          \\
                                        & \criticemoji                 &  $81.23$ & $71.24$          & $90.86$          &$85.24$           & $\textbf{87.96}$& $\textbf{70.54}$&   $45.15( 1.6e-14) $  \\
    \hline
    \multirow{3}{*}{\piemoji(PALM-2, \criticemoji(PALM-2))}& $f(x) = 1$&  $83.33$ & $50.0$        & $100.0$           & $83.33$          & $90.91$         & $ 50.0$         &   $ --(  --)$          \\
                                        & \criticemoji                 &  $84.49$ & $68.93$        & $93.9$            & $87.83$          & $\textbf{91.81}$& $\textbf{70.6}$&   $43.6(1.7e-15) $ \\
    \hline
    \multirow{3}{*}{\piemoji(PALM-2, \criticemoji(DePlot, PALM-2))}& $f(x) = 1$ &$89.4$ & $50.0$  & $100.0$          &$89.4$             & $94.41$         & $ 50.0 $        &   $ -- ( --)$          \\
                                        & \criticemoji                  &$92.38$ & $66.76$           & $99.26$          &$92.73$           & $\textbf{95.89}$& $\textbf{72.53}$&   $51.01(2.1e -21) $ \\
    \hline 
\end{tabular}
}
\caption{Evaluating \metric{} (\criticemoji) with a $0.75$ threshold against human labels on Chart-To-Text, we contrast with a no-op baseline ($f(x) = 1$) and report sentence binary classifier metrics.
For \pipeline (\piemoji), we generate $10$ candidates at temperature $T=0.7$, and \metric (\criticemoji) is computed with $T=0.3$ over 8 samples.}
\label{tab:results_Human_annotation} 
\end{table*}

\paragraph{Impact of critic model size.}%
We compare  in \Cref{tab:chats-critic-size} different LLMs to implement the entailment component of \metric{}.
We evaluate the performance of the models using the SciCap reference human annotation set and DePlot as a de-rendering model. We additionally study the entailment quality factoring out the de-rendering step by providing the original gold tables in SciCap and TabFact datasets.

As shown in the table, model size is a critical factor to improve \metric{} overall quality. In SciCap using DePlot respectively gold tables, we  see a $10.6$ respectively $12.6$ points increase on accuracy by using the small model compared to selecting all sentences (f(x)=1) and $11.3$ respectively $8.6$ increase when switching from small to large models. We observe the same behavior in TabFact with $4.2$ increase from small to large.

\begin{table}
\centering
\resizebox{1.\linewidth}{!}{
\begin{tabular}{llccc}
\toprule
    \textbf{Dataset}
    & \textbf{Sentence Selection metric}    & \textbf{Accuracy} & \textbf{F1}      &\textbf{AUC}      \\
    \hline
    \multirow{5}{*}{\rotatebox[origin=c]{90}{\parbox{2.3cm}{Statista \\ Reference}}}& $f(x) = 1$       &  $60.0$                   & $75.0$            & $ 50.0$           \\
                                        & \criticemoji(DePlot, PALM-2(S))       &     $70.67$               & $76.6$           &       $71.72$             \\
                                        & \criticemoji(DePlot, PALM-2(L))       &  $\textbf{82.0}$          & $\textbf{86.01}$  & $\textbf{81.56}$  \\
    \cline{2-5}
                                        & \criticemoji(PALM-2(S))               &  $72.67$                   & $77.09$           & $72.08$           \\ 
                                        & \criticemoji(PALM-2(L))               &  $\textbf{81.33}$          & $\textbf{86.0}$  & $\textbf{81.67}$  \\
    \hline 
  \multirow{3}{*}{\rotatebox[origin=c]{90}{TabFact}}           & $f(x) = 1$                            &  $50.32$           & $66.95$          & $ 50.0$           \\
                                        & \criticemoji(PALM-2(S))                &  $81.37$           & $79.45$          & $81.93$          \\
                                        & \criticemoji(PALM-2(L))               &  $\textbf{87.19}$  & $\textbf{87.23}$ & $\textbf{87.19}$  \\
                                        & \citet{eisenschlos-etal-2020-understanding} & $81.0$ & - & - \\
                                        & \citet{liu2022tapex} & $84.2$ & - & - \\
    \hline 
\end{tabular}
}
\caption{Comparing different critic models for \metric accuracy. For the L model, we use a threshold of 0.75 for SciCap and 0.5 for TabFact. For the S model, we use a threshold of 0.5 for both sets.
We also include two state-of-the-art finetuned models on the TabFact training set for reference.
We experimented with applying these trained models to Statista as well but the accuracy was very low due to poor generalization.}
\label{tab:chats-critic-size} 
\end{table}

\subsection{Metrics correlation to human ratings}
We investigate the correlation of the reference-based metrics to human ratings and compare it to \metric{}. 
Since these metrics are applied on the summary level, we extract the human entailment rating per summary: if any sentence is not entailed, the entire summary is refuted. 
PARENT~\citep{dhingra-etal-2019-handling} uses also the table information on top of the summary.
To thoroughly assess \metric{}, we report the correlation on summary level.

Additionally, we study the p-value and Pearson coefficient in \Cref{tab:results_Human_annotation_summaries}. To observe a possible correlation, reference-based metrics require optimizing for the entailment threshold (reported in the \Cref{app:results_Human_annotation_summaries_thresholds}).
Even accounting for that aspect, %
most of the reference-based metrics fail at providing a p-value that is statistically significant to identify a correlation (less than $0.05$). The majority of the metrics have a Pearson coefficient lower than $0.30$, indicating a small correlation. However, these metrics are less reliable than \metric, as these values are obtained by optimizing the threshold and the curve is not smooth; a deviation of $0.1$ in the threshold reduces the Pearson coefficient dramatically and increases the p-value. The results reported in the table, further confirm that our metric is more reliable and has a higher correlation with respect to the reference-based metrics. 
We additionally report the precision and recall curves for all metrics in \Cref{app:precision_and_recall_metrics}.

\subsection{Evaluation of the \pipeline{} pipeline}
In the second experimental setup, we compare in \Cref{tab:nlg-llm_results} different models to solve the chart-to-summary task on three data collections. 
We show that adding \pipeline{} improves any of the presented generative models on \metric{}. Additionally, it increases BLEURT-20 by around 1 point for all the data collections.
The best generative model is PALM-2. \pipeline(PALM-2) consistently reaches between $93\%$ and $96\%$ of \metric{}.
For more details, models and metrics results see \Cref{app:all-nlg-llm_results}.

\begin{table*}
\centering
\resizebox{0.96
\linewidth}{!}{
\begin{tabular}{lc|ccc}
    \toprule
    \textbf{Data collection}                & \textbf{$\metric_{summary}$}& \textbf{BLEURT-20} & \textbf{BLEU}      & \textbf{PARENT} \\ %
    \midrule
    Reference                                       &  $ 62.4 (1.6e-07)$        & $ 17.03 (2.0 e-1)$  & $nan$              & $24.2(6.7e-02)$\\ %
    PALM-2                                         &  $ 40.56 (1.5e-03)$       & $ 22.42 (9.0 e-02)$ & $ 14.92 (2.6e-01)$ & $ -29.57 (2.4e-02) $\\
    \piemoji(PALM-2, \criticemoji(PALM-2))          &  $ 50.27 (5.7e-05)$       & $ 25.59 (5.2e-02) $ & $ 34.16 (8.6e-03)$ & $ 28.33 (3.1e-02) $ \\
    \piemoji(PALM-2, \criticemoji(DePlot, PALM-2))    &  $ 51.97 (6.9e-04)$       & $ 45.68 (3.4e-03) $ & $26.92 (9.7e-01)$  & $23.62 (1.4e-01)$\\
    \bottomrule
\end{tabular}
}
\caption{Comparing models on \metric performance (i.e. \criticemoji{} column), instantiated with PALM-2 and DePlot for chart de-rendering.
\pipeline (i.e. rows with \piemoji{}) uses \metric configured in the same way. We report SciCap (First sentence) split for brevity.
Full results and additional evaluations are in \Cref{tab_app:all-nlg-llm_results}.} %
\label{tab:results_Human_annotation_summaries} 
\end{table*}

\begin{table}
\vspace{-1ex}
\centering
\resizebox{1.0\linewidth}{!}{
\begin{tabular}{clccc}%
   \toprule
   \textbf{Dataset}                        &  \textbf{Model}          & \textbf{\criticemoji} & \textbf{BLEURT} & \textbf{BLEU}\\
   \midrule
                                           & \citet{chen2023pali} PaLI-17B (res. 588) &$0.49$     &$0.49$   &$40.95$    \\ 
    Statista                               & DePlot+FLAN-T5           & $ 0.66$         &  $0.55$           &$42.5$ \\
                                           & PALM-2                   & $ 0.89$         &  $0.44 $          &$ 14.8$ \\
                                           & \piemoji~(DePlot+FLAN-T5)& $0.76 $         &  $\mathbf{0.57}$  &$\mathbf{43.1}$\\
                                           & \piemoji~(PALM-2)        & $\mathbf{0.96}$ &  $ 0.45$           &$13.34 $\\
    \hline
                                           & \citet{liu-etal-2023-matcha} MATCHA &--     &--                  &$12.2$    \\ 
                                           & \citet{chen2023pali} PaLI-17B (res. 588) &$0.35$     &$0.49$   &$13.93$    \\ 
    Pew                                    & DePlot+FLAN-T5           & $0.33$          &  $\mathbf{0.5}$     &$\mathbf{15.33}$  \\
                                           & PALM-2                   & $0.87$          &  $0.47$           &$8.83$ \\
                                           & \piemoji~(DePlot+FLAN-T5)& $0.41$          &$\mathbf{0.5}$              &$15.09$\\
                                           & \piemoji~(PALM-2)        & $\mathbf{0.95}$ &$0.48$            &$9.18$ \\
    \midrule
                                           &\citet{horawalavithana2023scitune} M4C-Captioner           &--& -- &$6.4$ \\
                                           & \citet{chen2023pali} PaLI-17B (res. 588) &$0.41$& $0.3$          &$11.05$    \\ 
    SciCap                                 & DePlot+FLAN-T5           & $0.34$          &  $0.29$           &$15.12$  \\
    {\small(First sentence)}                                    %
                                           & PALM-2                   &  $0.84$         &  $0.3$            &$0.94$ \\
                                           & \piemoji~(DePlot+FLAN-T5)& $0.48$          &  $0.3$            &$\mathbf{15.53}$ \\
                                           & \piemoji~(PALM-2)        & $\mathbf{0.93}$ &  $\mathbf{0.31}$   &$0.76$\\
    \bottomrule
\end{tabular}
}
\caption{Comparing models on \metric performance (i.e. \criticemoji{} column), instantiated with PALM-2 and DePlot for chart de-rendering.
\pipeline (i.e. rows with \piemoji{}) uses \metric configured in the same way. We report SciCap (First sentence) split for brevity.
Full results and additional evaluations are in \Cref{tab_app:all-nlg-llm_results}.} %
\label{tab:nlg-llm_results} 
\end{table}

\section{Analysis}
\paragraph{Ablation study (\pipeline{} 4 stages)} We report a performance study of the four stages of \pipeline, as depicted in \Cref{fig:ChaTS-Pi}, in \Cref{tab:results_stage_ablation}.
Droppings sentences in Stage 2 increases F1 by $1.9$ points compared to Stage 1.
Ranking with \metric{} without repair shows $8.3$ points compared to Stage 1 and $6.4$ to Stage 2.
Dropping the sentences of the top ranked summary increase F1 by $1.3$ reaching $95.69\%$ compared to using the top ranked summary.

\begin{table}[!ht]
\centering
\resizebox{0.85\linewidth}{!}{
\begin{tabular}{llcc}
    \toprule
    \multicolumn{2}{l}{\textbf{Stage name}}   & \textbf{F1} & \textbf{AUC} \\
    \midrule
    S1 & Summary generation                   & $86.03$ & $ 50.0$ \\
    S2 & Drop unentailed sentences            & $87.96$ & $ 70.54$ \\
    S3 & Summary scoring                      & $94.41$ & $50.0 $ \\
    S4 & \quad $\xhookrightarrow{}$ Filtering & $\textbf{95.69}$ & $\textbf{72.53}$ \\
    \hline 
\end{tabular}
}
\caption{Binary classification F1 and AUC of \pipeline{}'s different stages, on the PALM-2 annotation set. For an overview of the stages refer to \Cref{fig:ChaTS-Pi}.}
\label{tab:results_stage_ablation} 
\end{table}

We ablated the impact of DePlot on \metric{}, using the original tables as a baseline. The findings are detailed in \Cref{tab:results_gold_table_vs_deplot}. Given that DePlot's extracted tables may include missing or inaccurate data, we anticipated a greater sentence drop in \metric{} with DePlot. Contrarily, the F1 remains consistent for the reference and even sees an increase in \pipeline{} sets. Upon examining specific instances, we discerned the primary reason as following: Some numbers in gold tables are ``overly precise'' (sometimes several digits after the decimal, making it hard for humans to distinguish). In contrast, DePlot always outputs a ``rounded''/lossy value, which is preferred by human raters over those using the ultra-precise numbers from the gold table.
Despite these observations, the overall difference remains marginal (less than 1 percentage point). This suggests that DePlot's performance is commendably accurate, even when juxtaposed with gold tables.

\begin{table}[!ht]
\centering
\resizebox{0.89\linewidth}{!}{
\begin{tabular}{llllllcc}
    \toprule
    \textbf{Annotation set} & \textbf{Table} & \textbf{F1} & \textbf{AUC} \\
    \midrule
    \multirow{2}{*}{Reference}                                  & Gold     & $86.0$ & $\textbf{81.67}$ \\
                                                                & DePlot   & $\textbf{86.01}$ & $ 81.56$ \\
    \hline
    \multirow{2}{*}{PALM-2}                                     & Gold     & $\textbf{88.19}$ & $\textbf{71.97}$ \\
                                                                & DePlot   & $87.96 $ & $70.54$         \\
    \hline
    \multirow{2}{*}{\piemoji(PALM-2, \criticemoji(PALM-2))}         & Gold     & $91.52$ & $70.14$ \\
                                                                & DePlot   & $\textbf{91.81}$ & $\textbf{70.6}$ \\
    \hline
    \multirow{2}{*}{\piemoji(PALM-2, \criticemoji(DePlot, PALM-2))} & Gold     & $95.19$ & $\textbf{78.23}$ \\
                                                                & DePlot   & $\textbf{95.89}$ & $72.53$ \\
    \bottomrule
\end{tabular}
}
\caption{We compare performance of \metric{} when using a deplotter (DePlot) vs. using gold tables.}
\label{tab:results_gold_table_vs_deplot}
\end{table}

\paragraph{Grammaticality} defined as the human ratings on grammatical errors (see \Cref{subsec:evaluation-setup}) on non dropped sentences and summaries is reported in \cref{tab:results_Human_annotation_importance_grammar}. When applying \metric, we see a constant sentence-level \emph{Grammaticality} for the \pipeline last stage --The quality is already at $98.6\%$, leaving little room for improvement-- and a consistent improvement over all other sets. 
As for summary-level \emph{Grammaticality ($S$)}, the story is more nuanced.
On the Reference set (i.e. $\sim~3$ sentences per summary),
the impact on \emph{Grammaticality ($S$)} is less prominent.
On the PALM-2 annotation set, which features longer and more complex highlights (i.e. $\sim~5$ sentences per summary), we can see a small drop of $-1.97\%$.
\pipeline{} last stage remains constant, showing the importance of ranking.

\paragraph{Relevance} defined as the percentage of relevant sentences among the selected ones is reported in Table~\ref{tab:results_Human_annotation_importance_grammar}. We see a performance drop on this metric, mainly due to the design of \metric. The relevant sentences usually feature a more complex structure. \metric{} tends to prioritize less complex sentences during the entailment verification stage, thus producing an overall drop in \emph{Relevance}. 
One such example occurs when multiple statistics and computations are included in a single sentence: \emph{``The nonstore retailers increased by 22.8\% in April 2020 compared to April 2019, whereas store retailers jumped 12\% points in the same period.''}.
This is the case for the Reference and the \pipeline ranking stage.

\begin{table}[!ht]
\centering
\resizebox{1.\linewidth}{!}{
\begin{tabular}{lccc}
\toprule
    \textbf{Annotation Set}         & \textbf{Gram.}  & \textbf{Gram. ($S$)}  & \textbf{Relevance}\\
    \hline
    Reference                       &   $84.0$          &    $82.76$                &  $\mathbf{43.33}$\\
    Drop unentailed sentences       &   $\mathbf{88.29}$&    $\mathbf{83.93}$       &  $41.44$\\
    \hline
    PALM-2 Summary generation              &   $88.12$         &    $87.93$                &  $68.97$\\
    Drop unentailed sentences       &   $90.0$          &    $85.96$                &  $\mathbf{69.09}$\\
    Summary scoring                 &   $\mathbf{98.68}$&    $\mathbf{92.98}$       &  $58.94$\\
    \quad $\xhookrightarrow{}$ Filtering& $98.61$       &    $\mathbf{92.98}$       &  $57.14$\\
\bottomrule
\end{tabular}
}
\caption{
Human annotation rates for grammaticality and relevance computed on non dropped sentences.
\emph{Grammaticality} (Gram.) is the \% of grammatically correct sentences; \emph{Grammaticality ($S$)} the \% of fully grammatically correct summaries; \emph{Relevance} the percentage of relevant sentences.
We report the reference, and different \pipeline{} stages using PALM-2. \metric{} provides general improvements in sentence level grammaticality, whereas the performance on relevance and summary level grammaticality are mixed, due to \metric{} design.
}
\label{tab:results_Human_annotation_importance_grammar} 
\vspace{-2ex}
\end{table}

\subsection{Multilingual generalization}

\begin{figure}[!ht]
    \centering
    \begin{tabular}{>{\arraybackslash}p{0.45\textwidth}}
        \midrule[1.2pt]
        
        \multicolumn{1}{c}{\includegraphics[width=0.8\linewidth]{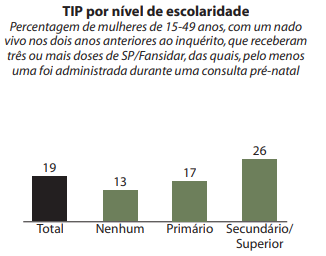}} \\
        \footnotesize\textbf{\textsc{\xmark \quad Summary:}} This statistic shows the percentage of women by education level . The percentage of women with a secondary or higher education is 26\% . The percentage of women with no education is 13\% . The percentage of women with primary education is 17\% . \\
        
        \midrule[1.2pt]
        \multicolumn{1}{c}{\includegraphics[width=0.8\linewidth]{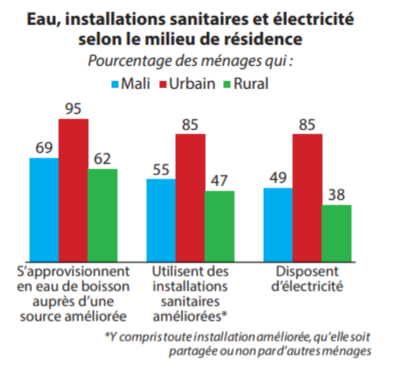}} \\
        \footnotesize\textbf{\textsc{\cmark \quad Summary:}} The statistics show the access to water, sanitation and electricity in Mali, by place of residence. In Mali, the access to improved drinking water sources is higher in urban areas (95\%) than in rural areas (62\%). The access to improved sanitation facilities is also higher in urban areas (85\%) than in rural areas (47\%). The access to electricity is higher in urban areas (85\%) than in rural areas (38\%). Overall, the access to basic services is lower in rural areas than in urban areas in Mali. \\
        
        \midrule[1.2pt]
    \end{tabular}
    
    \caption{Examples from TATA with \pipeline{} summaries using the demo and publicly accessible models.}
    \label{fig:i18n-examples}
    \vspace{-2ex}
\end{figure}

To evaluate the feasibility of our approach on \emph{internationalization} (i18n) datasets, we investigated a controlled setting using the TATA dataset~\cite{gehrmann2022tata}, a multilingual table-to-text dataset. 
We selected a localized image per language from the test set, including Portuguese, Arabic, French, Russian, and Swahili. 
Regrettably, the test set did not include images in Hausa, Yoruba, or Igbo. \Cref{fig:i18n-examples,fig:i18n-examples-second} include the images alongside the generated summaries.

Generally, the approach can provide an accurate summary if the deplotting component satisfies the following conditions: (a) no errors are introduced, or (b) a reasonable level of detail is provided. For Portuguese, the approach encounters failure case (b), as the deplotted table contains limited information (e.g., the subtext is not considered), resulting in a generic and factually deficient summary. Conversely, the error case (a) is observed for Arabic, where the information is extracted in an incorrect order, leading to factual inaccuracies.

When the table extraction is executed without errors, the summary generated is also accurate and reliable.%
However, it is acknowledged that there are specific instances where potential stylistic errors may manifest within the summary.
Such errors may include the reiteration of phrasal structures at the initiation of sentences or the potentially inappropriate integration of code-mixed information fragments.

\section{Related work}

\paragraph{Limits of reference-based metrics} have been explored for text summarization tasks and different solutions have been proposed to address them.
\citet{scialom-etal-2019-answers, scialom-etal-2021-questeval, fabbri-etal-2022-qafacteval} use question generation pipelines to evaluate the faithfulness of summaries without relying on references.
\citet{spice} study the use of intermediate dependency parse trees to rate natural image captions. 
For semi-structured data such as table-to-summary generation, PARENT~\cite{dhingra-etal-2019-handling} demonstrates the limitation of BLEU as it do not highlight the key knowledge from the table, whereas \citet{opitz-frank-2021-towards} focuses on generating text from abstract meaning representations. 
\citet{gehrmann2022tata} observed the poor correlation of BLEURT-20 to human ratings and proposed STATA, a learned metric using human annotation.

In this work we explore building a reference-free metric for chart summary that does not require human-annotated references. We show that our metric \metric has much higher correlation with human judgment than reference-based metrics such as BLEU.

\paragraph{Chart-to-summary generation} has recently gained relevance within multimodal NLP.
\citet{obeid-hoque-2020-chart} created the Chart-to-Text dataset, using charts extracted from Statista. \citet{kantharaj-etal-2022-chart} extended it with more examples from Statista and Pew Research.
Besides efforts on evaluation, multiple modeling methods have been proposed to reduce hallucinations. The approaches can be roughly divided into (1) pipeline-based methods which first extract chart components (e.g. data, title, axis, etc.) using OCR then leverage text-based models to further summarize the extracted information \citet{kantharaj-etal-2022-chart, choi2019visualizing}; (2) end-to-end models which directly input chart-attribute embeddings to Transformer-based models for enabling structured understanding of charts \cite{obeid-hoque-2020-chart}. 

In this work we explored both (1) and (2). The best approach \pipeline{} generally follows the idea of (1). Instead of relying on OCR we use a de-rendering model for extracting structured information in charts and we explore a self-critiquing pipeline with LLMs for the best quality chart summarization. 

\section{Conclusion}
In this paper, we tackle the chart-to-summary multimodal task, which has traditionally been challenging since it requires factual extraction and summarization of the insights presented in the image.
To measure the quality of a summary (especially faithfulness which has been overlooked by previous metrics), we present a reference-free metric called \metric{} \criticemoji{} for accurately and factually scoring chart-to-summary generation.
\metric{} obtains substantially higher correlations with human ratings compared to prior reference-based metrics. We additionally present \pipeline{} \piemoji, a self-critiquing pipeline to improve chart-to-summary generation. \pipeline{} leverages \metric{} scores to refine the output of any model by dropping unsupported sentences from the generated summaries and selecting the summary that maximizes fluency and \metric{}'scores. Compared with state-of-the-art baselines, \pipeline{} demonstrates stronger summarization quality across the board, achieving better scores for both the \metric which stresses faithfulness and also traditional metrics such as BLEURT.

\section*{Limitations}

In the following, we outline the limitations of our work to ensure transparency and inspire future research.
First, the chart domains we experimented with is limited to a few popular websites (e.g. Statista and Pew). This is due to the fact that existing academic chart-to-summary datasets only cover limited domains. However, to comprehensively evaluate the effectiveness of \metric{} and \pipeline{}, it is desirable to also evaluate our approaches in other chart domains such as infographics and scientific/financial charts.
Second, the \metric{} depends on a deplotter (image-to-text) model, specifically DePlot~\cite{liu-etal-2023-deplot}. DePlot has been trained on similar domains as the chart-to-summary datasets used in this work (e.g. Statista), and its performance may not generalize to other domains. The SciCap dataset evaluated in \Cref{tab:nlg-llm_results} provides some evidence of generalization. 
In future work, we plan to build out-of-domain evaluations to understand the impact of the deplotter's robustness better.
Third, we focused mostly on English chart summary in this work. We plan to also expand multilingual chart summary in future works and expanding our evaluation on the TaTa dataset \citep{gehrmann2022tata} as a test bed.

Another potential limitation is the use of \metric{} to evaluate \pipeline{}, although it is important to note that in order to mitigate the possibility for bias, we have cross checked with a variety of evaluation metrics and conducted ablation studies to analyze the individual components of our proposed solution. 
\pipeline{} has proven to also provide significant benefits when employing standard reference-based metrics. 
Comprehensive human studies have corroborated the usefulness of our results and we can confidently say that the improvement is consistent across the board.

We would also like to highlight the underlying risk of blindly trusting models to summarize content from an image accurately. Special care should be taken to verify outputs in accuracy-sensitive applications.

Despite its limitations, our work serves as an initial step in constructing reliable chart summarization evaluations and models. We hope future research can greatly benefit from this starting point.

\section*{Acknowledgments}
We would like to thank Massimo Nicosia,
Srini Narayanan, 
Yasemin Altun,
Averi Nowak,
Chenxi Pang,
Jonas Pfeiffer,
Mubashara Akhtar,
Parag Jain,
and the anonymous reviewers for their time, constructive feedback, useful
comments and suggestions about this work.

\bibliography{anthology,custom}
\bibliographystyle{style/acl_natbib}

\clearpage
\appendix
\section*{Appendix}

\section{Experimental Setup}
\subsection{Datasets}\label{app:datasets}
We use two popular chat-to-summary datasets for our experiments. The first one is Chart-to-Text~\cite{kantharaj-etal-2022-chart}, which can be found in \url{https://github.com/JasonObeid/Chart2Text}. The second one is SciCap~\cite{hsu-etal-2021-scicap-generating}, which is available at \url{https://github.com/
tingyaohsu/SciCap}. More details about the two datasets are introduced below.

\paragraph{Chart-To-Text} has mainly two sources: (i) Statista and (ii) Pew Research. (i) Statista is automatically extracted from an online platform that publishes charts in different topics including economics, market, and opinion; it is composed of 34,811 table, charts and summary triplets. (ii) Pew is automatically extracted then manually annotated from data-driven articles about social issues, public opinion and demographic trends; it is composed of 9,285 chart summary pairs.

\paragraph{SciCap~\cite{hsu-etal-2021-scicap-generating}} is a large-scale benchmark for figure-captioning. It is extracted from science arXiv papers published between 2010 and 2020 and contains more than 2 million figures. The figure-caption pairs are extracted using PDFFigures 2.0~\cite{clark2016ieee}, then an automatic figure type classifier is used to select graph plots. To be comparable to the work of \citet{hsu-etal-2021-scicap-generating}, we evaluate our model on the three subsets containing no sub-figures: \emph{First Sentence collection} including $133,543$ figures, \emph{Single-Sentence Caption} collection containing $94,110$ figures and \emph{Caption with No More than $100$ Words} composed of $131,319$ figures.

\subsection{De-rendering}
\label{app:Deplot}
We use DePlot~\citep{liu-etal-2023-deplot} model in all our experiments. The model code and checkpoint are available at \url{https://github.com/google-research/google-research/tree/master/deplot}. We use the GCS path to the base model \url{gs://deplot/models/base/deplot/v1} fine-tuned to solve the chart-to-table task. We do not perform any additional training, and use the model as a pre-processing step to extract the tables from the chart.

\subsection{Baselines}
We report the state-of-the-art models BLEU scores as presented in their papers. To be able to compare their models to ours and compute our new metric, we fine-tune a PaLI \cite{chen2023pali} model that gives a comparable results in BLEU as the other models. We select PaLI (Pathways Language and Image model) as our method of choice, because it takes the image as input directly, without the need for pre-processing or any OCR model to extract metadata, which can be difficult to reproduce.
In our experiments, we use the larger $17B$ variant and fine-tune for $5k$ iterations with an image resolution of $588\times588$. The PaLI model is fine-tuned with $128$ GCP-TPUv4. We use a batch size of $256$ and max sequence length of $128$.

\subsection{Our models}
\subsubsection{DePlot+T5 and DePlot+Flan-T5}
\label{app:flan_t5}
We adapt T5~\cite{raffel2020exploring} and FLAN-T5~\cite{suresh2023semantic} models: T5 is available at \url{https://huggingface.co/t5-base} and FLAN-T5 is available at \url{https://huggingface.co/google/flan-t5-base}. We adapt both base models to the chart-to-summary task. We add a de-rendering model to extract the table form the chart and use it as input of the models. Additionally, table embeddings are added to enhance table structure understanding. We fine-tune both models for $220k$ with $16$ GCP-TPUv3 cores using a batch size of $128$ and a max sequence length of $128$.

\subsubsection{MatCha-DePLot+FLAN-T5}
We use in our experiments MatCha-DePlot+FLAN-T5, which is composed of a MatCha~\cite{liu-etal-2023-matcha} image understanding module coupled to a DePlot+FLAN-T5 model, both of which are base size. MatCha base is available at
\url{https://github.com/google-research/google-research/tree/master/deplot}. This model takes in input both the a chart image and its table content (i.e. obtained by invoking DePlot). This setup should allow capturing visual aspects that DePlot ignores in its de-rendering process. MatCha-DePlot+FLAN-T5 is fine-tuned for $220k$ training steps with $32$ GCP-TPUv3, $128$ batch size, $1024$ image length and a max sequence length of $128$.

\subsubsection{PALM-2}
\label{app:LLM-prompting}
In our experiments for summary generation we use PALM-2(L)~\cite{anil2023palm} with in-context learning. The prompt is displayed in Figure~\ref{fig_app:summary_gen_prompt}.

\begin{figure*}
    \centering
\begin{lstlisting}[frame=single,basicstyle=\linespread{0.8}\ttfamily\tiny,breaklines=true,inputencoding=utf8,postbreak=\mbox{\textcolor{red}{$\hookrightarrow$}\space}]
Use a neutral tone and decontextualize information when possible. Use markdown format and bullet points for clarity.

Timeseries or data frame to consider:
the title is L'Oréal S.A. - worldwide revenue by division from 2012 to 2019 ( in million euros )

data =
col : Year | Consumer products | Professional products | L'Oréal Luxe | Active Cosmetics | The Body Shop
row 1 : 2019 | 12748.2 | 3441.9 | 11019.8 | 2663.7 | -
row 2 : 2018 | 12032.2 | 3262.5 | 9367.2 | 2275.5 | -
row 3 : 2017 | 12118.7 | 3350.4 | 8471.7 | 2082.9 | -
row 4 : 2016 | 11993.4 | 3399.7 | 7662.4 | 1860.7 | 920.8
row 5 : 2015 | 11844.2 | 3399.7 | 7230.0 | 1816.3 | 967.2
row 6 : 2014 | 10767.5 | 3032.4 | 6197.9 | 1660.4 | 873.8
row 7 : 2013 | 10873.2 | 2973.8 | 5865.2 | 1576.3 | 835.8
row 8 : 2012 | 10713.2 | 3002.6 | 5568.1 | 1499.2 | 855.3


Describe the data frame or time series. Identify important values such as min, max, sum, peak, bottom, increase, drop and highlight important comparisons, trends and unexpected behaviour, using maximum 5 sentences *in total*. Only reference data points you see.

Answer: This statistic shows L'Oréal 's global revenue from 2012 to 2019 , by division .
In 2016 , the Body Shop division of L'Oréal generated approximately 920.8 million euros in revenue .
Between 2012 and 2019 , the consumer products , the Professional products and L'Oréal Luxe divisions reached the highest values in 2019 .
The reported sum of revenue in 2019 is approximately 30k including the consumer products , the Professional products and L'Oréal Luxe divisions .

------

Use a neutral tone and decontextualize information when possible. Use markdown format and bullet points for clarity.

Timeseries or data frame to consider:
the title is Quarterly average daily rate of hotels in Dallas in 2016 and 2017 ( in U.S. dollars )

data =
col : Quarter | 2016 | 2017
row 1 : Q4 | 164 | -
row 2 : Q3 | 163 | -
row 3 : Q2 | 167 | -
row 4 : Q1 | 169 | 170


Describe the data frame or time series. Identify important values such as min, max, sum, peak, bottom, increase, drop and highlight important comparisons, trends and unexpected behaviour, using maximum 5 sentences *in total*. Only reference data points you see.

Answer: This statistic shows the quarterly average daily rate of hotels in Dallas in 2016 and 2017 .
From Q1 2016 to Q3 2016 the average daily rate of hotels in Dallas in the United States decreased a minimum value of 163 .
The rate started to increase from Q3 2016 .
In the first quarter (Q1) of 2017 , the rate reached the highest value 170 U.S. dollars .

------

Use a neutral tone and decontextualize information when possible. Use markdown format and bullet points for clarity.

Timeseries or data frame to consider:
the title is Cities with the largest number of community gardens per 10,000 residents in the United States in 2019


data =
col : city | Community gardens per 10,000 residents | Affected Community gardens per 10,000 residents
row 1 : Portland, OR | 3.8 | 1.9
row 2 : Madison, WI | 3.1 | 1.3
row 3 : St. Paul, MN | 3.8 | 1.9
row 4 : Orlando, FL | 2.4 | 1.3
row 5 : Washington, DC | 3.2 | 1.3
row 6 : Seattle, WA | 2.1 | 1.2

Describe the data frame or time series. Identify important values such as min, max, sum, peak, bottom, increase, drop and highlight important comparisons, trends and unexpected behaviour, using maximum 5 sentences *in total*. Only reference data points you see.

Answer: The statistics display the Cities with the largest number of community gardens per 10,000 residents in the United States in 2019 .
The statistics show that some cities are home to more community gardens than others .
In 2019 , both Portland , OR and St. Paul , MN had the largest number with 3.8 community gardens per 10,000 residents .
In 2019 , Seattle, WA had the lowest number with 2.1 community gardens per 10,000 residents .
The average values of Community gardens per 10,000 residents is 3.06 .
\end{lstlisting}
    \caption{PALM 3-shots prompting for summary generation}
    \label{fig_app:summary_gen_prompt}
\end{figure*}

\subsubsection{Critic model for \metric{}}
\label{app:prompting}
We use PALM-2~\cite{anil2023palm} as a critic model for \metric{}. Prompting is crucial for the interpretability of the entailment results. PaLM-2 outputs a text to refute or entail the claim. Following \citet{wei2022chain}, we use Chain-of-thought prompting to emphasize the reasoning before making the decision on the claim. More precisely we use 2 shots prompting for the critic models as shown in Figure~\ref{fig_app:chart2text_prompt}.
We use the same prompting for the large and small PALM-2 models.
The small model is available at \url{https://cloud.google.com/vertex-ai/docs/generative-ai/model-reference/text?hl=en}.

\begin{figure*}
    \centering
\begin{lstlisting}[frame=single,basicstyle=\linespread{0.8}\ttfamily\tiny,breaklines=true,postbreak=\mbox{\textcolor{red}{$\hookrightarrow$}\space}]
Read the table below regarding "Cities with the largest number of community gardens per 10,000 residents in the United States in 2019" to verify whether the provided claims are true or false.
col : city | Community gardens per 10,000 residents | Affected Community gardens per 10,000 residents
row 1 : Portland, OR | 3.8 | 1.9
row 2 : Madison, WI | 3.1 | 1.3
row 3 : St. Paul, MN | 3.8 | 1.9
row 4 : Orlando, FL | 2.4 | 1.3
row 5 : Washington, DC | 3.2 | 1.3
row 6 : Seattle, WA | 2.1 | 1.2

Claim: The statistics display the Cities with the largest number of community gardens per 10,000 residents in the United States in 2019 .
Answer: The title of the statistics is about the Cities with the largest number of community gardens per 10,000 residents in the United States in 2019 . Therefore, the claim is true .

Claim: The term 'community garden ' in the United States can mean a few different things .
Answer: The definition of community garden is not mentioned by the statistics . Therefore, the claim is false .

Claim:  For example , they can function as gathering places for the community and/or neighbors , however , they can also resemble the allotment gardens , often found in Europe , used by individuals and families .
Answer: The example is not mentioned by the statistics . Therefore, the claim is false .

Claim: Of course , some cities are home to more community gardens than others .
Answer: The statistics show that cities have different community gardens values . Therefore, the claim is true .

Claim: In 2019 , both Portland , OR and St. Paul , MN had the largest number with 3.8 community gardens per 10,000 residents .
Answer: Community gardens values are 3.8 > 3.2 > 3.1 > 2.4 > 2.1 . Portland , OR  Community gardens is 3.8 and St. Paul , MN  Community gardens is 3.8 . Community gardens per 10,000 residents highest number is 3.8 . Therefore, the claim is true .

Claim: In 2019 , Portland , OR and St. Paul , MN had the largest number with 1.9 community gardens per 10,000 residents .
Answer: Portland Community gardens is 3.8, OR and St. Paul, MN  Community gardens is 3.8 . 3.8 is not 1.9 . Therefore, the claim is false .

Claim: In 2019 , Washington, DC had the largest number .
Answer: Community gardens values are 3.8 > 3.2 > 3.1 > 2.4 > 2.1 . Washington, DC Community gardens is 3.2 . Both Portland , OR and St. Paul have the highest number 3.8 . Therefore, the claim is false .

Claim: In 2019  Portland, OR and Washington, DC had the largest number .
Answer: Community gardens values are 3.8 > 3.2 > 3.1 > 2.4 > 2.1 . Portland, OR Community gardens is 3.8 . St. Paul, MN is 3.8 . Washington, DC Community gardens is 3.2 . Both Portland , OR and St. Paul have the highest number 3.8 followed by Washington, DC 3.2 . Therefore, the claim is false .

Claim: In 2019  Washington, DC,  Portland, OR and St. Paul, MN had the largest number .
Answer: Community gardens values are 3.8 > 3.2 > 3.1 > 2.4 > 2.1 . Washington, DC Community gardens is 3.2 . Portland, OR Community gardens is 3.8 . St. Paul, MN is 3.8 . Both Portland , OR and St. Paul have the highest number 3.8 followed by Washington, DC 3.2 . Therefore, the claim is true .

Claim: In 2019 , Portland, OR had the largest number .
Answer:  Portland, OR Community gardens is 3.8 . St. Paul, MN Community gardens is also 3.8 . Both Portland, OR and St. Paul, had the largest number . Therefore, the claim is false .

Claim: In 2019 , Portland , OR and St. Paul , MN had the lowes number of affected Community gardens per 10,000 residents .
Answer: Affected Community gardens per 10,000 residents values are 1.9 > 1.3 > 1.2 . Both Portland , OR and St. Paul have Affected Community gardens is 1.9 . Community gardens per 10,000 residents lowes number is 1.2 . Therefore, the claim is false .

Claim: the average values of Community gardens per 10,000 residents is 3.06 .
Answer: The sum of Community gardens is 3.8 + 3.1 + 3.8 + 2.4 + 3.2 + 2.1 = 18.400000000000002 . There are 6 cities. The average is 18.40 / 6 = 3.0666666666666664 . Therefore, the claim is true .

Claim: the median values of Community gardens per 10,000 residents is 3.15 .
Answer: The ordered values of Community gardens are 3.8 , 3.8 , 3.2, 3.1 , 2.4 , 2.1 . There are 6 cities. The median refers to the values at positions 2 and 3 (6 / 2) . The median is 3.1 + 3.2  / 2 = 3.15 . Therefore, the claim is true .

------

Read the table below regarding "Number of households in Denmark from 2018 to 2020 by type of households
" to verify whether the provided claims are true or false.

col : Year | Married couple | Single women | Single men | Other couples | Other households including more than 1 family | Children below 18 years not living with parents
row 1 : 2020 | 934630 | 691059 | 527750 | 341985 | 231680 | 1028
row 2 : 2019 | 932591 | 682152 | 518279 | 338268 | 233819 | 962
row 3 : 2018 | 932254 | 676468 | 512552 | 335335 | 230875 | 988

Claim: Between 2018 and 2020 , the amount of households increased from 40702 to roughly 2.7 million .
Answer: The number of Married couple households in 2018 was 932254 + 676468 + 512552 + 335335 + 230875 + 988 = 2688472 and not 40702 . Therefore, the claim is false .

Claim: The number of households in Denmark increased by over 10 thousand in the period from 2019 to 2020 Answer .
Answer: The sum number of households in 2020 was 934630 + 691059 + 527750 + 341985 + 231680 + 1028 = 2728132 and in 2019 was 932591 + 682152 + 518279 + 338268 + 233819 + 962 = 2706071. The difference 2728132 - 2706071 = 22061 . 22061 is higher than 10000 . Therefore, the claim is true .

Claim: It reached its peak in 2020 .
Answer: The sum number of households in 2020 was 934630 + 691059 + 527750 + 341985 + 231680 + 1028 = 2728132. The sum number of households in 2019 was 932591 + 682152 + 518279 + 338268 + 233819 + 962 = 2706071. The sum number of households in 2018 932254 + 676468 + 512552 + 335335 + 230875 + 988 = 2688472 . 2728132 is higher than 2706071 and 2688472. The highest sum is in 2020. Therefore, the claim is true .

Claim: In 2019 it reached roughly 2.7 million .
Answer: The sum in 2019 is 932591 + 682152 + 518279 + 338268 + 233819 + 962 = 2706071. 2706071 is roughly 2.7 million. Therefore, the claim is false .
\end{lstlisting}
    \caption{PALM 2-shots prompting for \metric.}
    \label{fig_app:chart2text_prompt}
\end{figure*}

\section{Results}

\subsection{Annotation framework}
\label{app:annotation-framework}
\begin{figure}
    \centering
    \scalebox{1.07}{
    \includegraphics[width=0.9\linewidth]{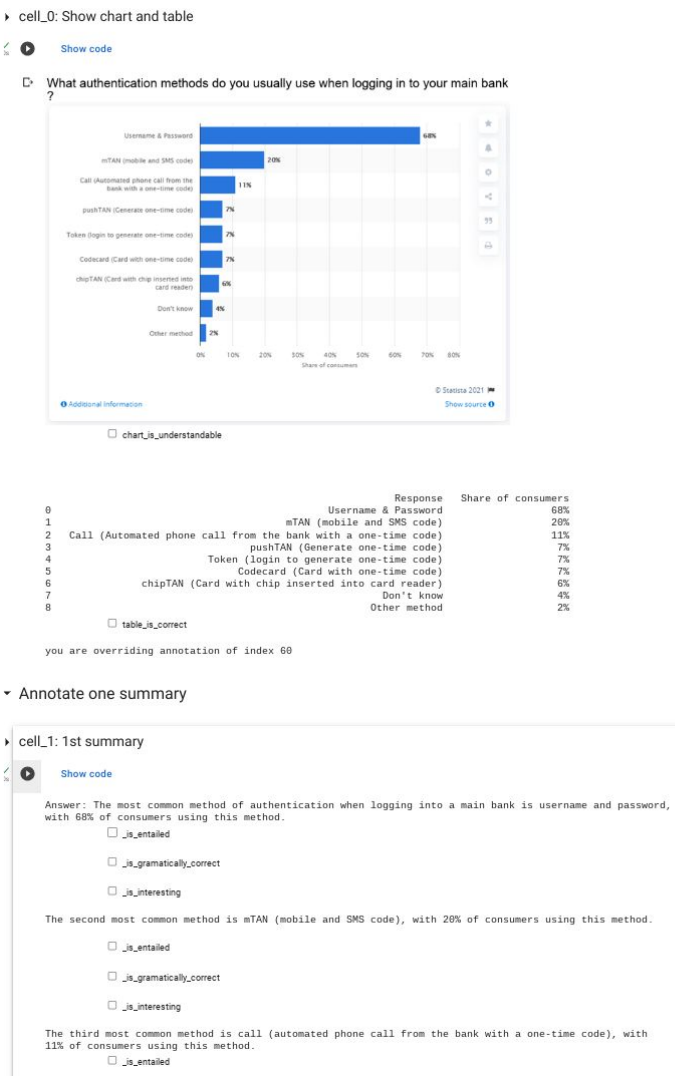}}
    \vspace{-2ex}
    \caption{Annotation system example}
    \label{fig_app:annotation-example}
    \vspace{-1.0ex}
\end{figure}

\Cref{fig_app:annotation-example} contains a screenshot of the annotation framework used to collect human ratings.

\subsection{Annotation guidelines}
\label{app:annotation_guidelines}
We provided to the raters the following annotation guidelines:

\begin{enumerate}
    \item \textbf{is\_interesting} highlights an important insight from the chart such as min / max / avg value or comparison.
    
    The \textbf{title} copy is \textbf{not interesting}.
    
    If the sentence is \textbf{not entailed or grammatically not correct} but highlights important info please select \textbf{is\_interesting}.
    
    \item \textbf{cleaned\_summary\_is\_grammatically} \textbf{\_correct =} grammar and fluency. Here the critic \textbf{model drops some sentences}. Please focus on the \textbf{fluency of the paragraph}.
    
    \item \textbf{Entailed =} you do not need additional info: \textbf{using the chart only}, be able to \textbf{extract the text}. (look at the chart the table can help you but not considered as ground truth.)
    
    If the prediction is equal to the title it is entailed you can consider it as not interesting.
    
    Please make sure that the meaning of the sentence does \textbf{not} add \textbf{additional info} about the chart.
    
    Examples:
    
    \begin{enumerate}
        \item The chart is about kids enrolled in kindergarten and nursery. The sentence contains: kids aged from 3 to 5. The title or the chart dose not refer to the age. \textbf{This adds a condition on the conducted study not referred in the title or the chart.} We considered it \textbf{not entailed}.\\

        \item If the sentence contains a \textbf{general knowledge} such as definitions:
        \begin{itemize}
            \item if \textbf{you know} that the definition \textbf{is correct} select \textbf{is\_entailed}
            \item if \textbf{you know} that it is \textbf{wrong} or \textbf{do not know} please select \textbf{not entailed}.
        \end{itemize}
    \end{enumerate}
    
    \item \textbf{Approximate numbers} is allowed up to \textbf{2 digits after the decimal.}
    
    Example: exact number in the chart between 2000 and 3000.
    \begin{itemize}
        \item Text\_1: "... around 2.51k" is \textbf{entailed}.
        \item Text\_2: "... around 2.5123k" is \textbf{not entailed}.
    \end{itemize}

    \item \textbf{grammatically\_correct =} look at grammar errors / fluency / repetition. Punctuation only if \textbf{it changes the meaning of the sentence}. Small errors are acceptable.
    
    Example: forget a letter/ invert letters / forget punctuation.
\end{enumerate}
\subsection{Correlation to human ratings}
\subsubsection{Pearson’s coefficient and p-value}
\label{app:results_Human_annotation_summaries_thresholds}
We extract the summary level human annotation as following; when a summary contains at least one unsupported sentence it is considered as unfaithful. As a result, the human summary annotation is binary while all the summary metrics BLEU, BLEURT-20, ChaTS-Critic are continuous. Without the use of a threshold to binarize them, the results of the p-value and Pearson coefficients are extremely uninformative. We choose the best possible threshold for each of the metrics in order to compare them in a fair way. In other words, we evaluate the classification task by selecting a threshold.
In the case of giving 2 binary vectors to compute the Pearson coefficient we have  Pearson = Spearman = Phi (standardized Chi-square). Table~\ref{tab_app:results_Human_annotation_summaries_thresholds} reports the different thresholds used to measure the p-value and Pearson’s coefficient in Table~\ref{tab:results_Human_annotation_summaries}.
\begin{table*}
\centering
\resizebox{1.\linewidth}{!}{
\begin{tabular}{lc|ccc}
    \toprule
    \textbf{Data collection}               &\textbf{$\metric_{summary}$}& \textbf{BLEURT-20 } & \textbf{BLEU}     & \textbf{PARENT} \\
    \midrule
    Reference                                                   &  $ 0.9$                   & $ 0.9$              & $nan$             & $0.79$\\ %
    PALM-2                                                     &  $ 0.9$                   & $ 0.4$              & $ 0.04$           & $ 0.3$\\
    \pipeline(PALM-2, \metric(PALM-2))                         &  $ 0.9$                   & $ 0.57$             & $ 0.13$           & $ 0.16$\\
    \pipeline(PALM-2, \metric(DePlot, PALM-2))                  &  $ 0.9$                   & $0.37$              & $ 0.16$           & $0.16$\\
    \bottomrule
\end{tabular}
}
\caption{The thresholds used to report the values in Table~\ref{tab:results_Human_annotation_summaries} were selected as follows. For all metrics except for \metric{}, we looked for the best threshold that maximized first the chance of observing a lower p-value and then a higher person coefficient. A constant threshold was considered for all sets when using \metric{}.}
\label{tab_app:results_Human_annotation_summaries_thresholds} 
\end{table*}

\subsubsection{Precision and Recall curves}
\label{app:precision_and_recall_metrics}
\begin{figure}
    \centering
    \includegraphics[width=0.7\linewidth]{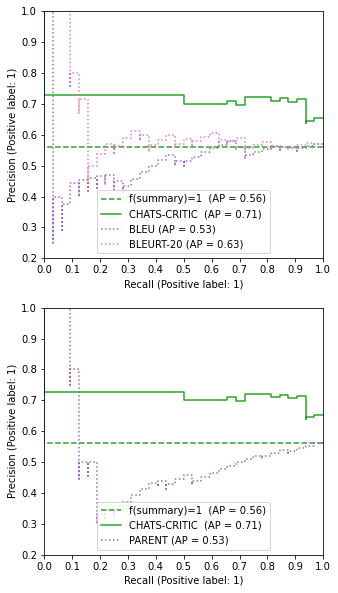}
    \caption{Human evaluation results for summary level to study the correlation of reference based metric to human ratings. Precision and Recall curve is displayed for the different metrics computed on Statista with summaries generated by PALM-2. The reported \metric{} Precision and Recall refers to the summary level.}
    \label{fig:results_Human_annotation_metrics} 
\end{figure}

Figure~\ref{fig:results_Human_annotation_metrics} shows the correlation of different metrics with human ratings by reporting Precision and Recall on the predicted summaries generated by PALM-2 compared to the original reference. A good correlation would display a continuously decreasing step function allowing to trade-off between Precision and Recall at a given threshold level. The \metric{} summary scores curve shows that it is a better classifier compared to all other metrics.

\subsection{\pipeline{} pipeline evaluation}
\label{app:Evaluation_of_the_summary_repair_pipeline}
We report supplementary experiments and baselines in Table~\ref{tab_app:all-nlg-llm_results}, alongside additional metrics. We report \metric{} and \pipeline{} using DePlot as a de-rendering model and if the original table is provided we add an extra row to ablate the effect of DePlot.
We use the following model checkpoint for BLEURT computation: \url{https://storage.googleapis.com/bleurt-oss-21/BLEURT-20.zip}.
\label{app:all-nlg-llm_results}
\begin{table*}
\centering
\resizebox{1.\linewidth}{!}{
\begin{tabular}{l|l|l|c|cc|c}
\toprule
   \textbf{Dataset}          & \textbf{Inputs}              &  \textbf{Model}          &\textbf{\metric} & \textbf{BLEURT-20} & \textbf{BLEU}             & \textbf{PARENT} \\
\midrule

    \multirow{19}{*}{Chart-To-Text (Statista)}&               & \citet{kantharaj-etal-2022-chart} TAB-T5 + (pretrained-pew)&--&$0.15$&$ 37.32$         & --\\               
                             &                              & TAB-T5                   & $0.55$          &  $0.53 $           &$ 40.48$     & $0.16 $ \\
                             &                              & TAB-FLAN-T5              & $0.67$          &  $0.56 $           &$ 41.48$     & $0.32 $ \\
                             &      original table          & PALM-2                   & $0.9$           &  $0.42 $           &$ 13.2$      & $0.52 $ \\
                             &      title                   & \piemoji(TAB-T5)        & $0.67$          &  $0.54 $           &$ 41.45$   & $0.16 $\\
                             &                              & \piemoji(TAB-FLAN-T5)   & $0.76$          &  $\mathbf{0.56}$   &$\mathbf{42.52}$& $0.32 $ \\
                             &                              & \piemoji(PALM-2)        & $\textbf{0.94}$ &  $0.45 $           &$ 12.43$     & $\mathbf{0.65}$ \\
    \cline{2-3}
                             &    original table           & MATCHA-TAB-FLAN-T5       & $0.68 $         &  $ 0.52$           &$38.57$       & $0.2 $\\
                             &     title + image            & \piemoji(MATCHA-TAB-FLAN-T5) & $0.71 $    &  $0.53 $           &$37.94 $     & $0.25 $ \\
   \cline{2-7}
                             &     OCR   table              & \citet{kantharaj-etal-2022-chart} OCR-T5    &--&$0.10$           &$35.29$     & -- \\ 
    \cline{2-3}
                             &    image                     & \citet{liu-etal-2023-matcha} MATCHA &--     &--                  &$39.4 $     & -- \\ 
                             &    image                    & \citet{chen2023pali} PaLI-17B (res. 588) &$0.49$      &$0.49$     &$40.95$   & -- \\ 
     \cline{2-3}
                             &                              & DePLot+T5                & $0.54$          &  $0.54$            &$41.83$      & $0.15 $  \\
                             &                              & DePLot+FLAN-T5           & $ 0.66$         &  $0.55$            &$42.5$       & $ 0.19$ \\
                             &     DePLot table             & PALM-2                   & $ 0.89$         &  $0.44 $           &$ 14.8$      & $0.32 $ \\
                             &     title                    & \piemoji(DePLot+T5)     & $ 0.66$         &  $ 0.56$           &$ 42.67$     & $ 0.15$\\
                             &                              & \piemoji(DePLot+FLAN-T5)& $0.76 $         &  $\mathbf{0.57}$   &$\mathbf{43.1}$ & $0.15 $ \\
                             &                              & \piemoji(PALM-2)        & $\mathbf{0.96}$ &  $ 0.45$           &$13.34 $     & $\mathbf{0.32}$\\
    \cline{2-3}
                             &     DePLot table             & MATCHA-DePLot+FLAN-T5    & $0.7 $          &  $ 0.54$           &$ 37.24$     & $0.25 $\\
                             &     title + image            & \piemoji(MATCHA-DePLot+FLAN-T5) & $0.79 $ &  $0.55 $           &$39.24 $     & $0.25 $ \\
    \hline\hline
    \multirow{10}{*}{Chart-To-Text (Pew)}& OCR   table         & \citet{kantharaj-etal-2022-chart} OCR-T5 &--&$-0.35$             &$10.49$         & --                 \\ 
    \cline{2-3}            
                             &    image                     & \citet{liu-etal-2023-matcha} MATCHA &--     &--                  &$12.2 $         & --  \\ 
                             &                              & \citet{chen2023pali} PaLI-17B (res. 588) &$0.35$     &$0.49$   &$13.93$     & --  \\ 
    \cline{2-3}    
                             &                              & DePLot+T5                & $0.27$          &  $0.49$            &$12.06$      & $0.04$ \\
                             &                              & DePLot+FLAN-T5           & $0.33$          &  $0.5$             &$15.33$      & $0.06$ \\
                             &     DePLot table             & PALM-2                   & $0.87$          &  $0.47$            &$8.83$       & $\mathbf{0.2}$ \\
                             &      title                   & \piemoji(DePLot+T5)     & $0.34$          &$0.5$               &$14.9$       & $0.04$ \\
                             &                              & \piemoji(DePLot+FLAN-T5)& $0.41$          &$0.5$               &$15.09$      & $0.06$ \\
                             &                              & \piemoji(PALM-2)        & $\mathbf{0.95}$ &$0.48$              &$9.18$       & $\mathbf{0.2}$ \\
    \cline{2-3}
                             &     DePLot table             & MATCHA-DePLot+FLAN-T5    &  $0.36$         &$\mathbf{0.51}$     &$\mathbf{15.41}$& $0$ \\
                             &      title                   & \piemoji(MATCHA-DePLot+FLAN-T5)&$0.42$    &$\mathbf{0.51}$     &$15.27$      & $0$ \\
    \hline\hline

    SciCap                   & SciTune info                 & \citet{horawalavithana2023scitune} LLaMA-SciTune (13B,CTOM) &--& -- &$5$         & -- \\
                             &                               & \citet{horawalavithana2023scitune} M4C-Captioner           &--& -- &$6.4$       & -- \\
    \hline\hline
    
    \multirow{10}{*}{SciCap (First Sentence)}& image           & \citet{hsu-etal-2021-scicap-generating} CNN+LSTM(vision only)  &--& -- &$2.19$         & -- \\
               & & \citet{chen2023pali} PaLI-17B (res. 588) &$0.41$& $0.3$          &$11.05$       & -- \\ 
    \cline{2-3}
                             &\multirow{6}{*}{DePLot table} &DePLot+T5                 & $0.3$           &  $0.28$            &$15.27$      & $0$  \\
                             &                              & DePLot+FLAN-T5           & $0.34$          &  $0.29$            &$15.12$      & $0.18$ \\
                             &                              & PALM-2                   &  $0.84$         &  $0.3$             &$0.94$       & $0.36$\\
                             &                              & \piemoji(DePLot+T5)     & $0.44$          &  $0.29$            &$15.2$       & $0.22$\\
                             &                              & \piemoji(DePLot+FLAN-T5)& $0.48$          &  $0.3$             &$\mathbf{15.53}$ & $0.18$\\
                             &                              & \piemoji(PALM-2)        & $\mathbf{0.93}$ &  $\mathbf{0.31}$   &$0.76$       & $\mathbf{0.36}$ \\
    \cline{2-3}
                             &      DePLot-table            & MATCHA-DePLot+FLAN-T5    & $0.36$          &  $0.29$            &$14.89$      & $0.2$\\
                             &      image                   & \piemoji(MATCHA-DePLot+FLAN-T5) &$0.49$   &  $0.3$             &$14.19$      & $0.49$ \\
   \hline\hline
   \multirow{9}{*}{SciCap (Single-Sent Caption)} & Text          & \citet{hsu-etal-2021-scicap-generating} CNN+LSTM(Text only)  &--& -- &$2.12$         & --  \\
    \cline{2-3}
                            & \multirow{6}{*}{DePLot table} & DePLot+T5                &$0.34$           &  $0.28$            &$13.27$       & $0$\\
                             &                              & DePLot+FLAN-T5           & $0.38$          &  $0.3$             &$15.28$      & $0.37$\\
                             &                              & PALM-2                   & $0.84$          &  $0.31$            &$0.69$       & $0.35$\\
                             &                              & \piemoji(DePLot+T5)     & $0.52$          &  $0.3$             &$15.72$      & $0$\\
                             &                              & \piemoji(DePLot+FLAN-T5)& $0.53$          &  $\mathbf{0.32}$   &$\mathbf{18}$ & $0.35$\\
                             &                              & \piemoji(PALM-2)        & $\mathbf{0.93}$ &  $\mathbf{0.32}$   &$0.61$       & $\mathbf{0.42}$\\
    \cline{2-3}
                             &      DePLot-table            & MATCHA-DePLot+FLAN-T5    &$0.37$           &  $0.3$             &$15.33$      & $0.23$ \\
                             &      image                   & \piemoji(MATCHA-DePLot+FLAN-T5)&$0.51$    &  $\mathbf{0.32}$   &$16.96$      & $0.23$ \\
  \hline\hline
   \multirow{9}{*}{SciCap (Caption w/ <=100 words)}  & Text  & \citet{hsu-etal-2021-scicap-generating} CNN+LSTM(Text only)  &--& -- &$1.72$         & --\\
    \cline{2-3}
                             & \multirow{6}{*}{DePLot table}&DePLot+T5                 & $0.31$          &  $0.28$            &$14.52$      & $0.15$\\
                             &                              & DePLot+FLAN-T5           & $0.35$          &  $0.29$            &$15.71$      & $0.17$\\
                             &                              & PALM-2                   & $0.82$          &  $0.3$             &$0.81$       & $0.41$\\
                             &                              & \piemoji(DePLot+T5)     & $0.45$          &  $0.29$            &$14.2$       & $0.15$\\
                             &                              & \piemoji(DePLot+FLAN-T5)& $0.48$          &  $0.3$             &$15.51$      & $0.17$\\
                             &                              & \piemoji(PALM-2)        & $\mathbf{0.93}$ &  $\mathbf{0.32}$   &$0.64$       & $0.46$\\
    \cline{2-3}
                             &      DePLot-table            & MATCHA-DePLot+FLAN-T5    & $0.34$          &  $0.29$            &$15.90$      & $\mathbf{0.48}$\\
                             &      image                   & \piemoji(MATCHA-DePLot+FLAN-T5)&$0.46$    &  $0.30$            &$\mathbf{16.16}$ & $\mathbf{0.48}$ \\
\bottomrule
\end{tabular}
}
\caption{Comparing different models on \metric{} performance. \metric{} refers to \metric{}(PALM-2) using the original table if it is provided in the input, else it refers to \metric{}(DePLot, PALM-2). Additionally, \pipeline{} \piemoji{} uses \metric{} following the same logic. The reported numbers for \citet{kantharaj-etal-2022-chart} uses BLEURT-128 base. For the our experiment we use BLEURT-20}
\label{tab_app:all-nlg-llm_results} 
\end{table*}

\subsection{Multilingual generalization}

\begin{figure}
    \centering
    \begin{tabular}{>{\arraybackslash}p{0.45\textwidth}}
        \midrule[1.2pt]

        \multicolumn{1}{c}{\includegraphics[width=0.8\linewidth]{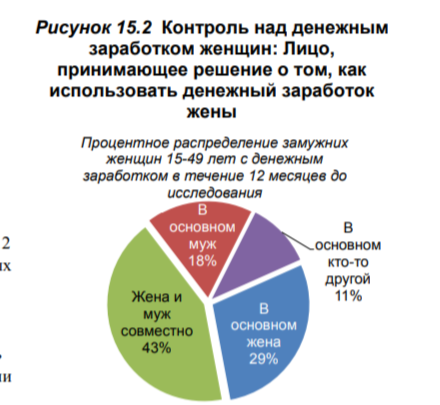}} \\
        \footnotesize\textbf{\textsc{\cmark \quad Summary:}} In the survey , 43\% of women reported that they and their partner made decisions about money together . 29\% of women reported that they made decisions about money on their own . 18\% of women reported that their partner made decisions about money on their own . 11\% of women reported that someone else made decisions about money . \\
        
        \midrule[1.2pt]

        \multicolumn{1}{c}{\includegraphics[width=0.8\linewidth]{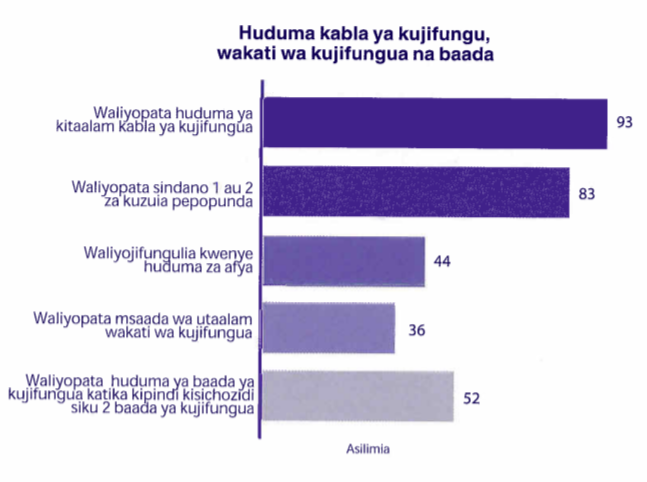}} \\
        \footnotesize\textbf{\textsc{\cmark \quad Summary:}} The statistics display the Huduma kabla ya kujifungua, wakati wa kujifungua na baada . The statistics show that 93\% of the people received professional care before giving birth . The statistics show that 83\% of the people received 1 or 2 injections to prevent tetanus . The statistics show that 44\% of the people gave birth in health services . The statistics show that 36\% of the people received expert assistance during childbirth . The statistics show that 52\% of the people received postpartum care within 2 days of giving birth . \\
        
        \midrule[1.2pt]
    \end{tabular}
    
    \caption{Additional i18n examples from TATA alongside with the summaries produced by \pipeline{} instantiated as \piemoji(Gemini, \criticemoji(Gemini)).}
    \label{fig:i18n-examples-second}
\end{figure}

\Cref{fig:i18n-examples-second} contains additional i18n examples from the TATA dataset~\cite{gehrmann2022tata}. Aside from a few stylistic quirks and redundancies, both summaries are accurate and faitfhful.

\end{document}